\documentclass{article}

\PassOptionsToPackage{numbers, compress, sort}{natbib}

\usepackage[final]{neurips_2021}

\usepackage[utf8]{inputenc} 
\usepackage[T1]{fontenc}    
\usepackage{hyperref}       
\usepackage{url}            
\usepackage{booktabs}       
\usepackage{amsfonts}       
\usepackage{nicefrac}       
\usepackage{microtype}      
\usepackage{xcolor}         

\usepackage{amsmath}
\usepackage{amssymb}
\usepackage{amsthm}
\usepackage{mathtools}
\usepackage{xspace}

\usepackage{amsmath,amsfonts,bm}

\def\eqref#1{equation~\ref{#1}}

\def\1{\bm{1}}

\def\vs{{\bm{s}}}

\def\mA{{\bm{A}}}
\def\mB{{\bm{B}}}

\def\mE{{\bm{E}}}

\def\mM{{\bm{M}}}

\def\mX{{\bm{X}}}

\DeclareMathAlphabet{\mathsfit}{\encodingdefault}{\sfdefault}{m}{sl}
\SetMathAlphabet{\mathsfit}{bold}{\encodingdefault}{\sfdefault}{bx}{n}

\newcommand{\E}{\mathbb{E}}

\newcommand{\R}{\mathbb{R}}

\DeclareMathOperator*{\argmax}{arg\,max}

\DeclareMathOperator*{\sargmax}{s\_arg\,max}

\newcommand{\bs}{\boldsymbol}

\newcommand{\loss}{\mathcal L}

\newcommand{\mMp}{\mM_{\text{pe}}}
\newcommand{\mMm}{\mM_{\text{mo}}}

\newcommand{\modelname}{\emph{ReMoto}\xspace}

\def\|#1|{\mathid{#1}}
\newcommand{\mathid}[1]{\ensuremath{\mathit{#1}}}
\def\<#1>{\codeid{#1}}
\protected\def\codeid#1{\ifmmode{\mbox{\smaller\ttfamily{#1}}}\else{\ttfamily
		#1}\fi}

\usepackage{algpseudocode}
\usepackage{algorithm}

\usepackage{arydshln}

\usepackage{enumitem} 

\usepackage{caption}
\usepackage{subcaption}
\usepackage{wrapfig}
\usepackage[mode=buildnew]{standalone}

\usepackage{CJKutf8}

\title{Structured Reordering for Modeling Latent Alignments in Sequence Transduction}
\author{
    Bailin Wang$^{1}$ \enskip Mirella Lapata$^{1}$ \enskip Ivan Titov$^{1,2}$ \\
    $^1$University of Edinburgh \enskip  $^2$University of Amsterdam \\
    {\tt bailin.wang@ed.ac.uk, \tt \{mlap, ititov\}@inf.ed.ac.uk}
}

\begin{document}

\maketitle

\begin{abstract}


Despite success in many domains, neural models struggle in settings
where train and test examples are drawn from different distributions.
In particular, in contrast to humans, conventional
sequence-to-sequence (seq2seq) models fail to generalize
systematically, i.e., interpret sentences representing novel
combinations of concepts (e.g., text segments) seen in
training. Traditional grammar formalisms excel in such settings by
implicitly encoding alignments between input and output segments, but
are hard to scale and maintain.  Instead of engineering a grammar, we
directly model segment-to-segment alignments as discrete structured
latent variables within a neural seq2seq model. To efficiently explore
the large space of alignments, we introduce a reorder-first
align-later framework whose central component is a neural reordering
module producing {\it separable} permutations. We present an efficient
dynamic programming algorithm performing exact marginal and MAP inference of
separable permutations, and, thus, enabling end-to-end differentiable 
training of our model.  The resulting seq2seq model exhibits better
systematic generalization than standard models on synthetic problems
and NLP tasks (i.e., semantic parsing and machine translation).
\end{abstract}

\section{Introduction}

\begin{wrapfigure}{r}{0.4\textwidth}
    \vspace{-1em}
    \centering
   \includegraphics[width=0.4\textwidth]{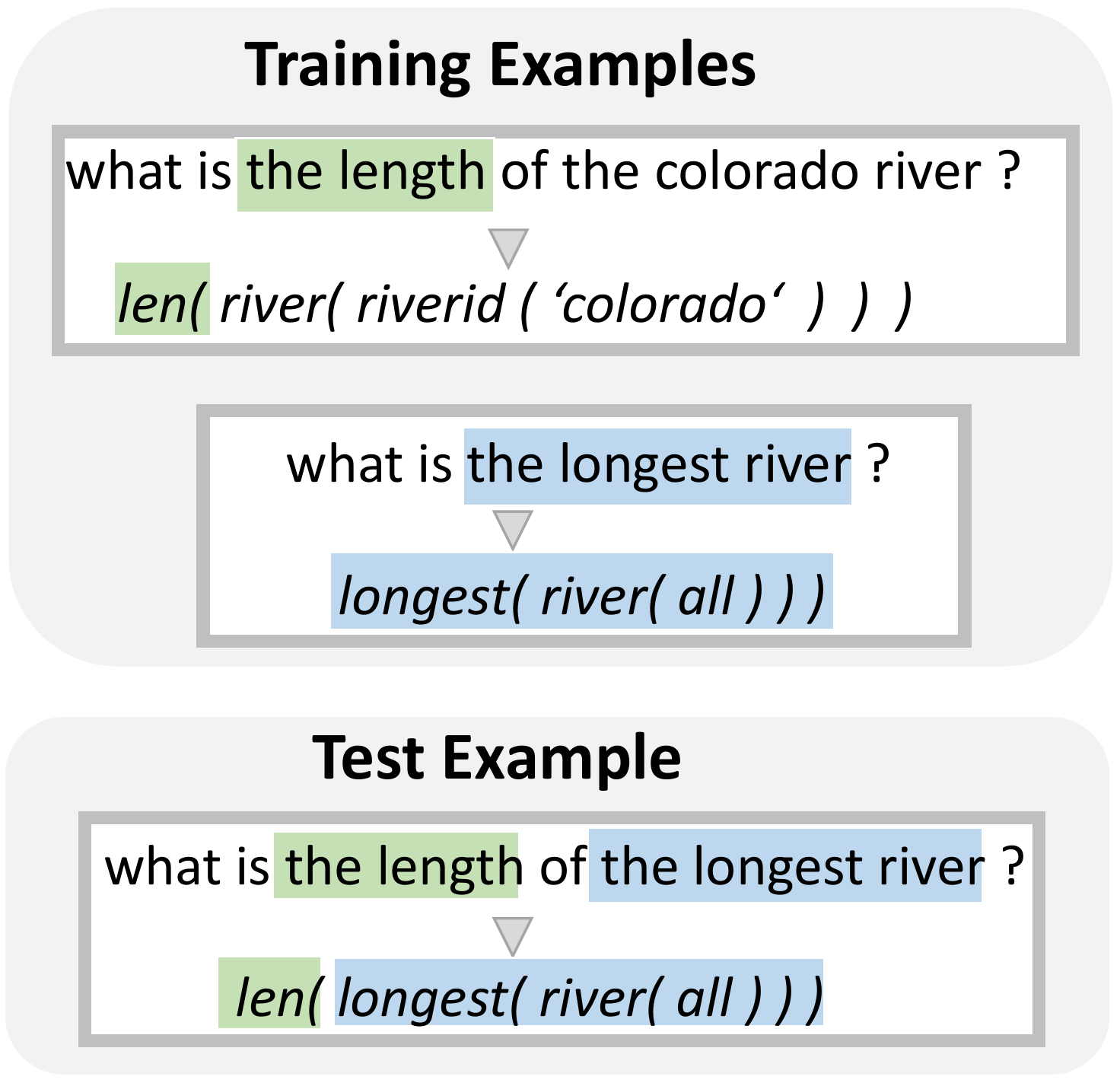} 
   \caption{A semantic parser needs to generalize to test examples which contain  segments from multiple training examples (shown in green and blue).}
   \label{fig:intro-example}
   \vspace{-1.5em}
\end{wrapfigure}
Recent advances in deep learning have led to major progress in many domains,  with neural models sometimes achieving or even surpassing human performance~\citep{wang2019superglue}. However, these methods often struggle in  out-of-distribution ({\it ood})  settings where train 
and test examples are drawn from different distributions. In particular, unlike humans, conventional sequence-to-sequence 
(seq2seq) models, widely used in natural language processing (NLP), fail to generalize 
{\it systematically}~\citep{bahdanau2018systematic,lake2018generalization,loula2018rearranging}, i.e.,~correctly interpret sentences representing novel combinations of concepts seen in training. 
 Our goal is to provide a mechanism for encouraging systematic generalization in seq2seq models.

To get an intuition about our method, consider the semantic parsing task shown in Figure~\ref{fig:intro-example}. A learner 
needs to map a natural language (NL) utterance to a program which can then be executed on a knowledge base. 
To process the test utterance, the learner needs to first decompose 
it into two segments  previously observed in training (shown in green and blue), and then 
combine their corresponding program fragments to create a new program.
Current seq2seq models fail in this systematic generalization setting~\cite{finegan-dollak-etal-2018-improving,keysers2019measuring}.
In contrast, traditional grammar formalisms decompose correspondences between utterances and programs into compositional mappings of substructures~\citep{steedman2000syntactic},  enabling grammar-based parsers to recombine rules acquired during training, as needed for systematic generalization. Grammars have proven essential in statistical semantic parsing in the pre-neural era~\cite{zettlemoyer2012learning,wong-mooney-2006-learning}, and have gained renewed interest now as a means of achieving systematic generalization~\cite{herzig2020span,shaw2020compositional}.  However, grammars are hard to create and maintain  (e.g.,~requiring grammar engineering or grammar induction stages) and  
do not scale well to NLP problems beyond semantic parsing (e.g.,~machine translation). 
In this work, we argue that the key property of grammar-based models, giving rise to their improved ood performance, is that a grammar implicitly encodes alignments between input and output segments. 
For example, in Figure~\ref{fig:intro-example}, the expected segment-level alignments are `\<the length $\rightarrow$ len>' and
`\<the longest river $\rightarrow$ longest(river(all))>'. The encoded alignments allow for \textit{explicit decomposition}
of input and output into segments, and \textit{consistent mapping} between input and output segments. In contrast, decision rules employed by 
conventional seq2seq models do not exhibit such properties. For example, recent work~\citep{goodwin2020probing} shows that primitive units such as words 
are usually inconsistently mapped across different contexts, preventing these models from generalizing primitive units to new contexts.
Instead of developing a full-fledged grammar-based method, we directly model segment-level alignments as structured latent variables. The resulting alignment-driven seq2seq model remains 
end-to-end differentiable, and, in principle, applicable to any sequence transduction problem.

Modeling segment-level 
alignments requires simultaneously inducing a segmentation of input and output sequences and discovering correspondences between the input and output segments.  
While segment-level alignments have been previously incorporated in neural models~\citep{yu2016online,wang2017sequence}, 
 to maintain tractability, these approaches support only monotonic alignments. The monotonicity assumption is reasonable for certain tasks (e.g.,~summarization), but it is generally overly restrictive  (e.g.,~consider semantic parsing and machine translation). 
To relax this assumption, we complement monotonic alignments 
with an extra reordering step. That is, we first permute the source sequence
so that segments within the reordered sequence can be aligned monotonically to segments of the target sequence. 
Coupling \emph{latent permutations} with monotonic alignments
dramatically increases the space of admissible segment alignments.

The space of general permutations is exceedingly large, so, to allow for efficient training, we restrict ourselves to \textit{separable} permutations~\citep{bose1998pattern}. We model separable permutations as hierarchical reordering of segments using {\it permutation trees}. This hierarchical way of modeling permutations reflects the hierarchical nature of language and hence is arguably more appropriate than `flat' alternatives~\cite{mena2018learning}. 
Interestingly, recent studies~\citep{steedman2020formal,stanojevic2018formal} demonstrated that separable permutations are sufficient for
capturing the variability of permutations in linguistic constructions across natural languages, providing further motivation for our modeling choice.


Simply marginalizing over all possible separable permutations remains intractable. Instead, inspired by recent work on modeling latent discrete structures~\citep{corro-titov-2019-learning,fu2020latent}, we introduce a continuous relaxation of the reordering problem. 
The key ingredients of the relaxation are two inference strategies: \textit{marginal inference}, which yields 
the expected permutation under a distribution; \textit{MAP inference}, which returns the most probable permutation.
In this work, we propose efficient dynamic programming algorithms to perform \emph{exact} marginal and MAP inference with separable permutations, 
resulting in effective differentiable neural modules producing relaxed separable permutations.
By plugging these modules into an existing module supporting monotonic segment alignments~\cite[][]{yu2016online}, 
we obtain end-to-end differentiable seq2seq models,  supporting non-monotonic segment-level alignments.

In summary, 
%
our contributions are:
\begin{itemize}[topsep=0pt,itemsep=2pt]
    \item A general seq2seq model for NLP tasks that accounts for latent non-monotonic segment-level alignments.
    \item Novel and efficient algorithms for exact marginal and MAP inference with separable permutations, allowing for end-to-end training using a 
     continuous relaxation.\footnote{Our code and data are available 
     at \url{https://github.com/berlino/tensor2struct-public}.
    }
    \item Experiments on synthetic problems and NLP tasks  (semantic parsing and machine translation) showing that modeling segment alignments is beneficial for systematic generalization.
\end{itemize}
\section{Background and Related Work}


\subsection{Systematic Generalization}

Human learners exhibit systematic generalization, which refers to their ability to generalize from training data  to novel situations.
This is possible due to the \textit{compositionality} of natural languages
-  to a large degree,  sentences are built using an inventory of primitive concepts and finite structure-building mechanisms~\citep{chomsky_aspects_1965}.
For example, if one understands `John loves the girl',
they should also understand `The girl loves John'~\citep{fodor1988connectionism}.
This is done by `knowing' the meaning of individual words and the grammatical principle of subject-verb-object composition. 
As pointed out by~\citet{goodwin2020probing}, systematicity entails that primitive units have consistent meaning across different contexts. 
In contrast, in seq2seq models, the representations of a word are highly influenced by  context (see experiments in \citet{lake2018generalization}). This is also consistent with the observation that seq2seq models tend to memorize large chunks rather than discover underlying compositional principles~\citep{hupkes_compositionality_2019}. The memorization of large sequences lets the model fit the training distribution but harms out-of-distribution generalization. 


\subsection{Discrete Alignments as Conditional Computation Graphs}



Latent discrete structures enable the incorporation of inductive biases into neural models and have been beneficial for a range of problems. 
 For example, 
input-dependent module layouts~\cite{andreas2016neural} or graphs~\cite{norcliffe2018learning} have been explored
in visual question answering. There is also a large body of work on inducing task-specific discrete representations (usually trees) for NL sentences \cite{yogatama2016learning,niculae2018sparsemap,havrylov-etal-2019-cooperative,corro-titov-2019-learning}. The trees are induced simultaneously with learning a model performing a computation relying on the tree (typically a  recursive neural network~\cite{socher2011semi}), while optimizing a task-specific loss. Given the role the structures play in these approaches -- i.e.,  defining the computation flow -- we can think of the structures as {\it conditional computation graphs}.



In this work, we induce discrete alignments as conditional computation graphs to guide seq2seq models.
Given a source sequence $x$ with $n$ tokens 
and a target sequence $y$ with $m$ tokens, we optimize the following objective:
\begin{align}
    \begin{split}
       \mX = \text{Encode}_{\theta} (x) \quad \quad \quad
       \loss_{\theta, \phi}(x,y)  = -\log \E_{p_{\phi}(\mM | \mX)} p_{\theta} (y| \mX, \mM) 
    \end{split}
    \label{eq:general_obj}
\end{align}
where $\text{Encode}$ is a function that embeds $x$ into $\mX \in \R^{n \times h}$ with $h$
being the hidden size, $\mM \in \{0, 1\}^{n \times m}$ is the alignment matrix between
input and output tokens. In this framework, alignments $\mM$ are separately predicted by $p_{\phi}(\mM | \mX)$  to guide 
the computation $p_{\theta} (y| \mX, \mM)$ that maps $x$ to $y$. The parameters of both model components ($\phi$ and $\theta$) are disjoint.

\paragraph*{Relation to Attention}
Standard encoder-decoder models~\citep{bahdanau2014neural}
rely on continuous attention weights i.e., $\mM[:, i] \in \triangle^{n-1}$ for each target token $1\leq i \leq m$. 
Discrete versions of attention (aka hard attention) have been studied in previous work~\citep{xu2015show,deng2018latent} and show 
superior performance in certain tasks. In the discrete case $\mM$ is a sequence of $m$ categorical random variables.
Though discrete, the hard attention only considers word-level alignments, i.e., assumes that each target token is aligned with a single source token.  This is a limiting assumption; for example,
in traditional statistical machine translation, word-based models (e.g., \citep{brown1993mathematics}) are known to achieve dramatically weaker results than phrase-based models (e.g., \citep{koehn-etal-2007-moses}).
In this work, we aim to bring the power of phrase-level (aka segment-level) alignments to neural seq2seq models.~\footnote{
One of our models (see Section~\ref{ssection:soft-reordering}) still has a flavor of standard continuous attention in that it approximates 
discrete alignments with continuous expectation. }
\section{Latent Segment Alignments via Separable Permutations}
\label{sec:reorder}

\begin{figure}[t]
    \vspace{-1em}
    \centering
   \includegraphics[width=.9\textwidth]{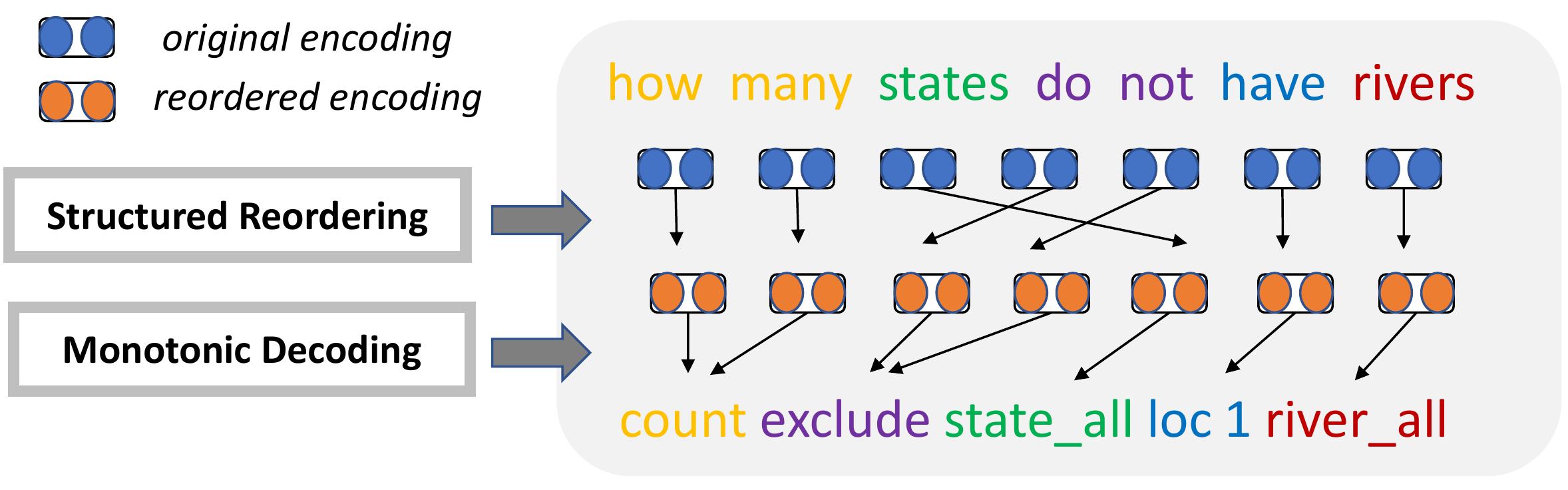} 
   \caption{The architecture of our seq2seq model for semantic parsing. 
   After encoding the input utterance, our model permutes the input representations using our
   reordering module. Then, the reordered encodings will be used 
   for decoding the output program in a monotonic manner. }
   \label{fig:arch}
   \vspace{-1em}
\end{figure}

%
Our method integrates a layer of segment-level alignments with a seq2seq model.
The architecture of our model is shown in Figure~\ref{fig:arch}. 
Central to this model is the alignment network, which decomposes the alignment problem into two stages: (i) input  reordering and (ii) monotonic alignment between 
the reordered sequence and the output.
%
%
%
%
Conceptually, we decompose the alignment matrix from Eq~\ref{eq:general_obj} into two parts:
\begin{equation}
    \mM = \mMp \mMm
\end{equation}
where $\mMp \in \R^{n \times n}$ is a permutation matrix, 
and $\mMm \in \R^{n \times m}$ represents monotonic alignments. 
With this conceptual decomposition, we can rewrite the objective in Eq~\ref{eq:general_obj}
as follows:
\begin{align}
    \begin{split}
       \loss_{\theta, \phi}(x,y) & = -\log \E_{p_{\phi}(\mMp |x)} 
       \E_{p_{\phi'}(\mMm|\mMp \mX)} p_{\theta} (y| \mMp \mX, \mMm)
    \end{split}
    \label{eq:general_obj_decomp}
\end{align}
where $\mMp \mX $ denotes the reordered representation.
With a slight abuse of notation, $\phi$ now denotes the parameters of the model generating 
permutations, and $\phi'$ denotes the parameters used to produce monotonic alignments.
Given the permutation matrix $\mMp$,
the second expectation $\E_{p_{\phi}(\mMm|\mMp \mX)} p_{\theta} (y| \mMp \mX, \mMm)$,
which we denote as $p_{\theta, \phi'}(y|\mMp\mX)$, can be handled by existing methods, such as 
SSNT~\citep{yu2016online} and SWAN~\citep{wang2017sequence}. 
In the rest of the paper, we choose SSNT as the module for handling monotonic alignment.\footnote{
In our initial experiments, we found that SWAN works as well as SSNT but is considerably slower.} 
We can rewrite the objective we optimize in the following compact form:
\begin{align}
    \begin{split}
       \loss_{\theta, \phi, \phi'}(x,y) & = -\log \E_{p_{\phi}(\mMp |x)}  p_{\theta, \phi'}(y|\mMp \mX)
    \end{split}
    \label{eq:general_obj_short}
\end{align}

\subsection{Structured Latent Reordering by Binary Permutation Trees}

Inspired by~\citet{steedman2020formal}, we restrict word reorderings to separable permutations.
Formally, separable permutations are defined in terms of binary permutation trees (aka separating trees~\citep{bose1998pattern}), i.e., if a permutation can be represented by a permutation tree, it is separable.
A binary permutation tree over a permutation of a sequence $1\dots n$ is a binary tree in which each node represents the ordering of a segment $i \dots j$; the children exhaustively split their parent into sub-segments $i \dots k$ and $k+1 \dots j$.
Each node has a binary label that decides whether the segment of the left child precedes that of the right child. 
\begin{wrapfigure}{r}{0.4\textwidth}
    \vspace{-10pt}
    \includegraphics[width=0.38 \textwidth]{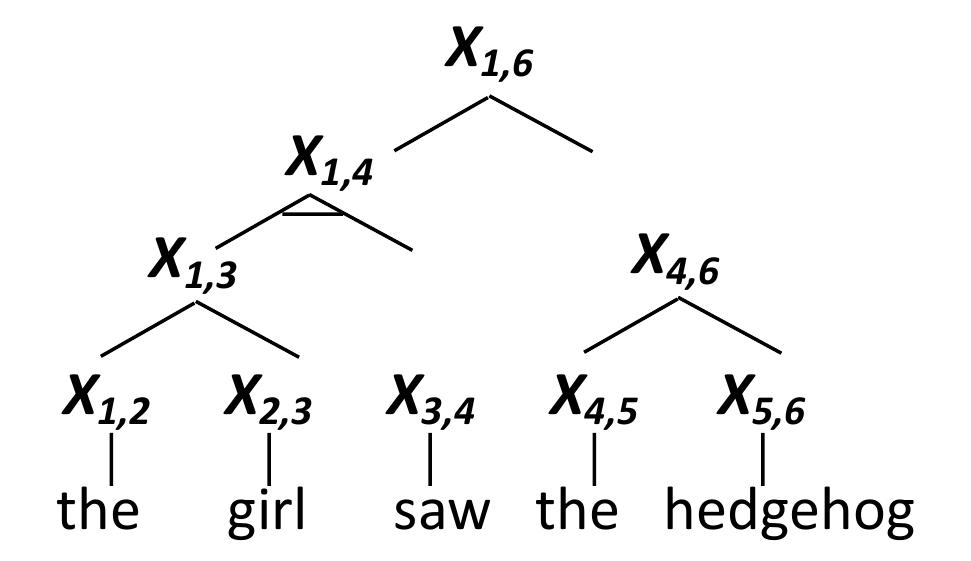}
    \caption{The tree represents the reordered sentences `saw the girl the hedgehog' where
    $\triangle, \wedge$ denotes \emph{Inverted} and \emph{Straight}, respectively.}
    \label{fig:btg-example}
    \vspace{-40pt}
\end{wrapfigure}
Bracketing transduction grammar~\cite[BTG,][]{wu1997stochastic}, which is proposed in the context of machine translation, is the corresponding context-free grammar to represent binary permutation trees.
Specifically, BTG has one non-terminal ($X$) and three anchored rules: 
\begin{description}[noitemsep]
    \item $\mathcal{S}_{i,j,k}$ \ : \quad  $X_i^k \xrightarrow{Straight} X_i^j X_{j}^k$
    \item $\mathcal{I}_{i,j,k}$ \ :  \quad  $X_i^k \xrightarrow{Inverted} X_i^j X_{j}^k$
    \item $\mathcal{T}_{i}$ \quad \ \ :  \quad $X_i^{i+1} \rightarrow x_i$
\end{description}
where $X_i^k$ is the anchored non-terminal covering the segment from $i$ to $k$ (excluding $k$).
The first two rules decide whether to keep or invert two segments when constructing a larger segment;
the last rule states that every word $x_i$ in an utterance is associated with a non-terminal $X_i^{i+1}$.
An example is shown in Figure~\ref{fig:btg-example}.
Through this example, we note that the first two rules only signify which segments to inverse; an additional process of interpreting 
the tree (i.e., performing actual actions of keeping or inverting segments) is needed to obtain the permutated sequence.
%
This hierarchical approach to generating separable permutations reflects the compositional nature of language, 
and, thus, appears more appealing than using `flat' alternatives~\cite{mena2018learning,grover2019stochastic,cuturi2019differentiable}. 
Moreover, with BTGs, we can incorporate segment-level
features to model separable permutations, and design tractable algorithms for learning and inference.

By assigning a score to each anchored rule using segment-level features, we obtain a distribution over all 
possible derivations, and use it to compute the objective in Eq~\ref{eq:general_obj_short}.
\begin{equation}
    p_\phi(D|x) = \frac{\prod_{R \in D} f_\phi(R)}{Z(x, \phi)}, \quad
    \loss_{\theta, \phi, \phi'}(x,y)  = -\log \E_{p_{\phi}(D |x)}  p_{\theta, \phi'}(y|\mMp^D \mX)
    \label{eq:obj_final}
\end{equation}
where $f_\phi$ is a score function assigning a (non-negative) 
weight to an anchored rule $R \in \{\mathcal{S}, \mathcal{I}, \mathcal{T} \}$, 
$Z(x, \phi) = \sum_{D'} \prod_{R \in D'} f_\phi(R)$ 
is the partition function, which can be computed using the 
inside algorithm, $\mMp^D$ is the permutation matrix corresponding 
to the derivation $D$.
BTG, along with the weight assigned for each rule, is 
a weighted context-free grammar (WCFG). 
In this WCFG, the weight is only normalized 
at the derivation level.
As we will see in Algorithm~\ref{algo:marginal-sampling}, we are interested in normalizing the weight of production rules 
and converting the WCFG to an 
equivalent PCFG following~\citet{smith2007weighted},
so that the probability of  a derivation can be computed as follows:
\begin{equation}
    p_\phi(D|x) = \prod_{R \in D} G_\phi(R)
    \label{eq:btg_pcfg}
\end{equation}
where $G_\phi(R)$ is the weight of the production rule $R$ under the transformed PCFG.
The details of the conversion are provided in the Appendix.

The challenge with optimizing the objective in Eq~\ref{eq:obj_final} is that 
the search space of possible derivations is exponential, making the estimation
of the gradients with respect to parameters of the reordering component ($\phi$) non-trivial.
We now present two differentiable surrogates we use. 

\subsection{Soft Reordering: Computing Marginal Permutations}
\label{ssection:soft-reordering}

\begin{wrapfigure}{r}{0.5\textwidth}
    \vspace{-6mm}
    \begin{minipage}{0.5\textwidth}
    \begin{algorithm}[H]
        \caption{Dynamic programming for computing marginals and differentiable sampling of permutation matrix wrt. a parameterized grammar}
        \textbf{Input:} $G_{\phi}(R)$: probability of an anchored rule $R$ \\
        \hspace*{2.7em} \textit{sampling}: whether perform sampling
        \begin{algorithmic}[1]
            \For{$i := 1$ \textbf{to} $n$} 
                \State  $\mE_i^{i+1} = \bs 1$ 
            \EndFor
            \For{$w := 2$ \textbf{to} $n$} \Comment width of spans
                \For{$i := 1$ \textbf{to} $n - w + 1$} 
                    \State $k := i + w$ 
                    \If{\textit{sampling}}
                        \State $\hat G_\phi(R) = \sargmax(G_\phi(R))$\label{algo:step-sampling} 
                    \Else  \Comment computing marginals
                        \State $\hat G_\phi(R) = G_\phi(R)$ 
                    \EndIf
                    \For{$j := i+1$ \textbf{to} $k-1$}
                       \State $\mE_i^k \mathrel{+}= \hat G_\phi(\mathcal{S}_{i,j,k}) (\mE_i^j \oplus \mE_j^k) $\label{algo:step-marginalize-1} 
                       \State $\mE_i^k \mathrel{+}= \hat G_\phi(\mathcal{I}_{i,j,k}) (\mE_i^j \ominus \mE_j^k) $\label{algo:step-marginalize-2} 
                    \EndFor
                \EndFor
            \EndFor
            \State \Return $\mE_1^{n+1}$
        \end{algorithmic}
        \label{algo:marginal-sampling}
    \end{algorithm}
\end{minipage}
\vspace{-8mm}
\end{wrapfigure}

The first strategy is to use the deterministic expectation of permutations to
softly reorder a sentence, analogous to the way standard attention approximates 
categorical random variables. Specifically, we use the following approximation:
\begin{align*}
    \mMp' &= \E_{p_{\phi}(D |x)} \mMp^D \\ 
    \loss_{\theta, \phi, \phi'}(x,y) & \approx -\log   p_{\theta, \phi'}(y|\mMp'\mX)
\end{align*}
where $\mMp'$ is the marginal permutation matrix, and it can be treated as structured attention~\citep{kim2017structured}.
Methods for performing marginal inference for anchored rules, i.e., computing the marginal distribution of production rules are well-known in NLP~\citep{manning1999foundations}.
However, we are interested in the marginal permutation matrix (or equivalently the expectation of the matrix components) as the matrix is the data structure that is ultimately used in our model. 
As a key contribution of this work, we propose an efficient algorithm to exactly compute the marginal permutation matrix using dynamic programming. 

In order to compute the marginal permutation matrix we need to marginalize over the exponentially many derivations of each permutation.
We propose to map a derivation of BTG into its corresponding permutation matrix in a recursive manner.
Specifically, we first associate word $i$ with an
identity permutation matrix $\mM_i^{i+1} = \bs 1$;
then we associate \emph{Straight} and \emph{Inverted} rules with direct $\oplus$ and skew $\ominus$ sums 
of permutation matrices, respectively:
\begin{equation*}
    \mA \oplus \mB = 
        \begin{bmatrix}
            \mA & \bs 0 \\
            \bs 0 & \mB
        \end{bmatrix}  
    \quad \quad
    \mA \ominus \mB = 
        \begin{bmatrix}
            \bs 0 & \mA \\
            \mB & \bs 0
        \end{bmatrix}
\end{equation*}
For example, the permutation matrix of the derivation tree
shown in Figure~\ref{fig:btg-example} can be obtained by:
\begin{equation}
  \mM_1^6 = \bigg( \big( (\mM_1^2 \oplus \mM_2^3) \ominus \mM_3^4 \big) \oplus (\mM_4^5 \oplus \mM_5^6) \bigg )
\end{equation}
Intuitively, the permutation matrix of long segments can be constructed by 
composing permutation matrices of short segments. Motivated by this, 
we propose a dynamic programming algorithm, which takes advantage of the observation that 
we can reuse the 
permutation matrices of short segments when 
computing permutation matrices of long segments, as shown in Algorithm~\ref{algo:marginal-sampling}. While the above equation is defined over discrete permutation matrices encoding a single derivation, the algorithm applies recursive rules to expected permutation matrices. 
Central to the algorithm is the following recursion:
\begin{equation}
   \mE_i^k = \sum_{i<j<k} G_\phi(\mathcal{S}_{i,j,k}) (\mE_i^j \oplus \mE_j^k) + G_\phi(\mathcal{I}_{i,j,k}) (\mE_i^j \ominus \mE_j^k)
\end{equation}
where $\mE_i^k$ is the expected permutation matrix for the segment from $i$ to $k$, $G_\phi(R)$ is 
the probability of employing the production rule $R$, defined in Eq~\ref{eq:btg_pcfg}.
Overall, Algorithm~\ref{algo:marginal-sampling} is a bottom-up method that constructs expected permutation matrices incrementally in 
Step~\ref{algo:step-marginalize-1} and \ref{algo:step-marginalize-2}, while relying on the probability of the associated
production rule. We prove the correctness of this algorithm by induction in the Appendix.

\subsection{Hard Reordering: Gumbel-Permutation by Differentiable Sampling}

During \textbf{inference}, for efficiency, it is convenient to rely on the most probable derivation $D'$ and its corresponding 
most probable $y$:
\begin{equation}
    \argmax_{y} p_{\theta, \phi'}(y|\mMp^{D'} \mX)
    \label{eq:obj_inference}
\end{equation}
where $D' = \argmax_D p_\phi(D|x)$.
The use of discrete permutations $\mMp^{D'}$ during inference and soft reorderings during training lead to a training-inference gap which may be problematic. 
Inspired by recent Gumbel-Softmax operator~\citep[][]{jang2016categorical,maddison2016concrete} that relaxes 
the sampling procedure of a categorical distribution using the Gumbel-Max trick, we propose a differentiable 
procedure to obtain an approximate sample $\mMp^D$ from $p(D|x)$. 
Concretely, the Gumbel-Softmax operator relaxes the perturb-and-MAP procedure~\citep{papandreou2011perturb}, where we add noises to probability
logits and then relax the MAP inference (i.e., $\argmax$ in the categorical case); we denote this operator as $\sargmax$.
In our structured case, we perturb the logits of the probabilities of production rules $G_\phi(R)$, and  relax 
the structured MAP inference for our problem.
Recall that $p(D|x)$ is converted to a PCFG, and MAP inference for PCFG is algorithmically similar 
to marginal inference. Intuitively, for each segment, instead of marginalizing over 
all possible production rules in marginal inference, we choose the one with the highest probability (i.e., a local MAP inference with categorical random variables) 
during MAP inference. By relaxing each local MAP inference with Gumbel-Softmax (Step~\ref{algo:step-sampling} of Algorithm~\ref{algo:marginal-sampling}), 
we obtain a differentiable sampling procedure.~\footnote{If we change $\sargmax$ with $\argmax$ in Step~\ref{algo:step-sampling} of 
Algorithm~\ref{algo:marginal-sampling}, we will obtain the algorithm for exact MAP inference. 
}
We choose Straight-Through Gumbel-Softmax so that the return of Algorithm~\ref{algo:marginal-sampling} is a discrete permutation matrix, and in this way we close
the training-inference gap faced by soft reordering.

\paragraph{Summary} We propose two efficient algorithms for computing marginals and obtaining samples of separable 
permutations with their distribution parameterized via BTG. In both algorithms, PCFG plays an important role of decomposing a global problem into sub-problems, which explains why we convert $p(D|x)$ into a PCFG in Eq~\ref{eq:btg_pcfg}.
Relying on the proposed algorithms, we present two relaxations of the discrete permutations that let us induce latent reorderings with end-to-end training. We refer to the resulting system
as \modelname, short 
for a seq2seq model with \underline{Re}ordered-then-\underline{Mo}no\underline{to}ne alignments.
Soft-\modelname and Hard-\modelname denote the versions which use soft marginal permutations and 
hard Gumbel permutations, respectively.

\paragraph{Segment-Level Alignments}
Segments are considered as the basic elements being manipulated in our reordering module.
Concretely, permutation matrices are constructed by hierarchically reordering input segments.
SSNT, which is the module on top of our reordering module for monotonically generating output, conceptually
also considers segments as basic elements. Intuitively, SSNT alternates between consuming an input segment 
and generating an output segment. 
Modeling segments provides a strong inductive bias, reflecting the intuition that sequence transduction in NLP can be largely accomplished by manipulations at the level of segments. In 
contrast, there is no explicit notion of segments in conventional seq2seq methods. 

However, different from our reordering module where segments are first-class objects during modeling, the alternating process of SSNT
is realized by a series of token-level decisions (e.g., whether to keep consuming the next input token). 
Thus, properties of segments (e.g., segment-level features) are not fully exploited in SSNT. 
In this sense, one potential way to further improve \modelname is to explore better alternatives to SSNT that can 
treat segments as first-class objects as well. We leave this direction for future work.

\paragraph{Reordering in Previous Work}
In traditional statistical machine translation (SMT), reorderings are typically handled by a distortion model~\citep[e.g.,][]{al2006distortion} in a pipeline manner.
\citet{neubig-etal-2012-inducing}, \citet{nakagawa-2015-efficient} and  \citet{stanojevic2015reordering} also use BTGs for modeling reorderings. 
 \citet{stanojevic2015reordering} 
 go beyond binarized grammars, showing how to support 5-ary branching permutation trees. 
Still, they assume the word alignments have been produced on a preprocessing step, using an alignment tool~\cite{och-ney-2003-systematic}. Relying on these alignments, they induce reorderings. Inversely, we rely on latent reordering to induce 
the underlying word and segment alignments.
%

Reordering modules have been previously used in neural models, and can be assigned to the following two categories. 
First, reordering components \citep{huang2017towards,chen2019neural} 
were proposed for neural machine translation. However, they are not structured or sufficiently constrained
in the sense that they may produce invalid reorderings (e.g., a word is likely to be moved to 
more than one new position). In contrast, our module is a principled way of dealing with latent reorderings.
Second, the generic permutations (i.e., one-to-one matchings or sorting), though having differentiable counterparts~\citep{mena2018learning,grover2019stochastic,cuturi2019differentiable},
do not suit our needs as they are defined in terms of tokens, rather than segments. 
For comparison, in our experiments, we design baselines that are based on Gumbel-Sinkhorn Network~\citep{mena2018learning},
which is used previously in NLP~(e.g., \cite{lyu-titov-2018-amr}).

\section{Experiments}
\label{sec:experiment}

First, we consider two diagnostic tasks where we can test the neural reordering module on its own.
Then we further assess our general seq2seq model \modelname on two real-world NLP tasks,
namely semantic parsing and machine translation.

\subsection{Diagnostic Tasks}
\label{sec:diagnostic}


\begin{table}
    \center
    \scalebox{0.9}{
    \begin{tabular}{l|l|l} 
        \toprule
        Dataset & Input & Output \\
        \midrule
        Arithmetic & $((1+9)*((7+8)/4))$ & $((19+)((78+)4/)*)$ \\ 
        SCAN-SP & jump twice after walk around left thrice & after (twice (jump), thrice(walk (around, left)))  \\
        GeoQuery & how many states do not have rivers ? & count(exclude(state(all), loc\_1(river(all))))  \\
        \bottomrule
       \end{tabular}
    }
    \caption{Examples of input-output pairs for parsing tasks.}
    \label{tab:example_data}
\vspace{-3mm}
\end{table}

\begin{table}
    \center
    \begin{tabular}{ l|cc|cc } 
        \toprule
        & \multicolumn{2}{c|}{\textbf{Arithmetic}} & \multicolumn{2}{c}{\textbf{SCAN-SP}} \\
        Model & \textsc{IID} & \textsc{Len} & \textsc{IID} & \textsc{Len} \\ 
        \hline
        Seq2Seq & 100.0 & 0.0 & 100.0 & 13.9  \\ 
        LSTM-based Tagging  & 100.0 & 20.6 & 100.0 & 57.7  \\
        Sinkhorn-Attention Tagging & 99.5 & 8.8 & 100.0 & 48.2 \\ 
        \hdashline
        Soft-\modelname  & 100.0 & \textbf{86.9} & 100.0 & 100.0  \\
        \quad \emph{- shared parameters}  & 100.0 & 40.9 & 100.0 & 100.0 \\ 
        Hard-\modelname  & 100.0 & 83.3 & 100.0 & 100.0  \\
        \bottomrule
       \end{tabular}
    \caption{Accuracy (\%) on the arithmetic and SCAN-SP tasks.}
    \label{tab:synthetic}
    \vspace{-6mm}
\end{table}

\paragraph{Arithmetic} 
We design a task of converting an arithmetic expression in infix format to the one in postfix format. 
An example is shown in Table~\ref{tab:example_data}. 
We create a synthetic dataset by sampling data from a PCFG.
In order to generalize, a system needs to learn how to manipulate internal sub-structures (i.e., segments) while respecting well-formedness constraints. 
This task can be solved by the shunting-yard algorithm but we are interested to see if neural networks can solve it and generalize ood by learning from raw infix-postfix pairs. 
For standard splits (\textsc{IID}), we randomly sample 
20k infix-postfix pairs whose nesting depth is set to be between 1 and 6; 
10k, 5k, 5k of these pairs are used as train, dev and test sets, respectively. 
To test systematic generalization, we create a Length split (\textsc{Len}) where 
training and dev examples remain the same as \textsc{IID} splits, but test examples 
have a nesting depth of 7. In this way, we test whether 
a system can generalize to unseen longer input.

\paragraph{SCAN-SP} We use the SCAN dataset~\cite{lake2018generalization}, which consists of simple English 
commands coupled with sequences of discrete actions. Here we use the semantic parsing version, 
 SCAN-SP~\cite{herzig2020span}, where the goal is to predict programs corresponding to the action sequences. An example is shown in~Table~\ref{tab:example_data}.
As in these experiments our goal is to test the reordering component alone, 
we 
remove parentheses and commas in programs. For example, the program \<after (twice (jump), thrice(walk (around, left)))  >
is converted to a sequence: \<after twice jump thrice walk around left>.
In this way, the resulting parentheses-free sequence can be viewed as a reordered sequence of the NL utterance `jump twice after walk around left thrice'.
The grammar of the programs is known so we can reconstruct the original program from 
the intermediate parentheses-free sequences using the grammar.
Apart from the standard split (\textsc{IID}, aka simple split~\citep{lake2018generalization}), 
we create a Length split (\textsc{Len}) where the training set contains NL utterances with a maximum
length 5, while utterances in  the dev and test sets have a minimum length of 6.\footnote{
Since we use the program form, the original length split~\citep{lake2018generalization}, which is based on the length of action sequence, is not very suitable in our experiments.}

\paragraph{Baselines and Results}
In both diagnostic tasks, we use \modelname with a trivial monotonic alignment matrix $\mMm$ (an identity matrix)
in Eq~\ref{eq:general_obj_decomp}. Essentially, \modelname becomes a sequence tagging model.
We consider three baselines: (1) vanilla Seq2Seq models with 
Luong attention~\cite{luong2015effective}; (2) an LSTM-based tagging model which
learns the reordering implicitly, and can be viewed as a version \modelname with a trivial $\mMp$
and $\mMm$; (3) Sinkhorn Attention that replaces the permutation matrix of Soft-\modelname  in Eq~\ref{eq:general_obj_short} 
by Gumbel-Sinkhorn networks~\cite{mena2018learning}. 

We report results by averaging over three runs in Table~\ref{tab:synthetic}.
In both datasets, almost all methods
achieve perfect accuracy in \textsc{IID} splits. However, baseline systems 
cannot generalize well to the challenging \textsc{LEN} splits. 
In contrast, our methods, both Soft-\modelname and Hard-\modelname,  perform
very well on \textsc{Len} splits, surpassing the best baseline system by large margins ($>40\%$).
The results indicate that \modelname, particularly its neural reordering 
module, has the right inductive bias to learn reorderings.
We also test a variant Soft-\modelname where parameters $\theta, \phi$ with shared input embeddings.
This variant does not generalize well to the \textsc{LEN} split on the arithmetic task, showing 
that it is beneficial to split models of the `syntax' (i.e., alignment) and `semantics', confirming what has been previously observed~\cite{havrylov-etal-2019-cooperative,russin2019compositional}. 

\subsection{Semantic Parsing}

Our second experiment is on semantic parsing where \modelname models 
the latent alignment between NL utterances and their corresponding programs.
We use GeoQuery dataset~\cite{zelle1996learning}
which contains 880 utterance-programs pairs. 
The programs are in variable-free form~\cite{kate2005learning}; 
an example is shown in Table \ref{tab:example_data}.~\footnote{
We use the varaible-free form, as opposed to other alternatives such lambda calculus, for two reasons:
1) variable-free programs have been commonly used in systematic generalization settings~\citep{herzig2020span,shaw2020compositional}, 
probably it is easier to construct generalization splits using this form; 2) the variable-free form is more suitable 
for modeling alignments since variables in programs usually make alignments hard to define.}
Similarly to SCAN-SP, we transform the programs into
parentheses-free form which have better structural correspondence with utterances.
Again, we can reconstruct the original programs 
based on the grammar. 
An example of such parentheses-free form is shown in Figure~\ref{fig:arch}.
Apart from the standard version, we also experiment with the Chinese 
and German versions of GeoQuery~\citep{jones-etal-2012-semantic,susanto2017semantic}.
Since different languages exhibit divergent word orders~\cite{steedman2020formal},
the results in the multilingual setting will tell us if 
our model can deal with this variability. 

In addition to standard \textsc{IID} splits, we create a \textsc{Len} split 
where the training examples have parentheses-free programs with  a maximum length 4;
the dev and test examples have programs with a  minimum length 5.
We also experiment with the \textsc{Temp} split~\cite{herzig2020span}
where training and test examples have programs with disjoint templates.

\begin{table}
    \center
    \scalebox{1.0}{
    \begin{tabular}{ l|ccc|ccc|ccc } 
        \toprule
        & \multicolumn{3}{|c|}{\textbf{EN}} & \multicolumn{3}{|c|}{\textbf{ZH}} & \multicolumn{3}{|c}{\textbf{DE}} \\
        Model & \textsc{IID} &  \textsc{Temp} & \textsc{Len} & \textsc{IID} & \textsc{Temp} & \textsc{Len}  & \textsc{IID}  & \textsc{Temp} & \textsc{Len} \\ 
        \cmidrule(lr){1-1}
        \cmidrule(lr){2-4}
        \cmidrule(lr){5-7}
        \cmidrule(lr){8-10}
        Seq2Seq & 75.7 & 38.8 & 21.8 & 72.5 & 25.4 & 19.8 & 56.1 & 18.8 & 15.2 \\ 
        Syntactic Attention~\citep{russin2019compositional} & 74.3 & 39.1 & 18.3 & 70.2 & 27.9 & 18.7 & 54.3 & 19.3 & 14.2  \\
        SSNT~\citep{yu2016online} & 75.3 & 38.7 & 19.1 & 71.6 & 23.8 & 17.8 & 55.2 & 19.8 & 14.1 \\
        \hdashline
        Soft-\modelname   & 74.5  & 39.3  & 19.8  & 73.4 & 30.3 & 17.3 & 55.8 & 19.5 & 13.4   \\
        Hard-\modelname   & 75.2  & \underline{43.2}  & \underline{23.2} & \underline {74.3} & \underline{45.7} & \underline{22.3} & 55.6 & \underline{22.3} & 16.6  \\
        \bottomrule
       \end{tabular}
    }
    \caption{Exact-match accuracy (\%) on three splits of the multilingual GeoQuery dataset. 
    Numbers underlined are significantly better than others 
    (p-value $\leq 0.05$ using the paired permutation test). 
    }
    \label{tab:parsing}
    \vspace{-6mm}
\end{table}

\paragraph{Baselines and Results}
Apart from conventional seq2seq models, for comparison, we also implemented the syntactic attention~\cite{russin2019compositional}. Our model \modelname is similar in spirit to the syntactic attention,  
`syntax' in their model (i.e., alignment) and `semantics' (i.e., producing the representation relying on the alignment) are separately modeled. 
In contrast to our structured mechanism for modeling alignments, their syntactic attention still relies on the 
conventional attention mechanism.
We also compare with SSNT, which can be viewed as an ablated version of \modelname by removing our reordering module.

Results are shown in Table~\ref{tab:parsing}. 
For the challenging \textsc{Temp} and \textsc{Len} splits, our best performing model Hard-\modelname
achieves consistently stronger performance than seq2seq, syntactic attention and SSNT. 
Thus, our model bridges the gap between conventional seq2seq models 
and specialized state-of-the-art grammar-based models~\cite{shaw2020compositional,herzig2020span}.\footnote{
NQG~\citep{shaw2020compositional} achieves 35.0\% in the English \textsc{LEN}, and 
SBSP~\citep{herzig2020span} (without lexicon) achieves 65.9\% in the English \textsc{TEMP} in execution accuracy. Both models are augmented with pre-trained representations (BERT).}

\subsection{Machine Translation}

Our final experiment is on small-scale machine translation tasks, where \modelname
models the latent alignments between parallel sentences from two 
different languages. To probe systematic generalization, we also 
create a \textsc{LEN} split for each language pair in addition to 
the standard \textsc{IID} splits.

\paragraph{English-Japanese}
We use the small en-ja dataset extracted from TANKA Corpus.
The original split (\textsc{IID}) has 50k/500/500 examples 
for train/dev/test with lengths 4-16 words.\footnote{\url{https://github.com/odashi/small_parallel_enja}} 
We create a \textsc{LEN} split where the English sentences of training examples 
have a maximum length 12 whereas the English sentences in dev/test have a minimum length 13. The \textsc{LEN} split has 50k/538/538 examples for train/dev/test, respectively.

\paragraph{Chinese-English}

We extract a subset from FBIS corpus (LDC2003E14) by filtering English sentences with length 4-30.
We randomly shuffle the resulting data to obtain an \textsc{IID} split which 
has 141k/3k/3k examples for train/dev/test, respectively. 
In addition, we create 
a \textsc{LEN} split where English sentences of training examples 
have a maximum length 29 
whereas the English sentences of dev/test examples have a length 30. The \textsc{LEN} split has 140k/4k/4k examples as train/dev/test sets respectively. 

\begin{wraptable}{r}{0.6\textwidth}
    \center
    \vspace{-5mm}
    \scalebox{1.0}{
    \begin{tabular}{ l|cc|cc } 
        \toprule
        & \multicolumn{2}{c|}{\textbf{EN-JA}} & \multicolumn{2}{c}{\textbf{ZH-EN}} \\
        & \multicolumn{1}{c}{\textsc{IID}} & \multicolumn{1}{c|}{\textsc{Len}} & \multicolumn{1}{c}{\textsc{IID}} & \multicolumn{1}{c}{\textsc{Len}} \\
        \midrule
        Seq2Seq & 35.6 &  25.3 &  21.4 &  18.1 \\ 
        SSNT~\citep{yu2016online}   & 36.3 &  26.5 &  20.5 &  17.3 \\
        Local Reordering~\citep{huang2017towards} & 36.0 &  27.1 &  21.8 &  17.8 \\ 
        \hdashline
        Soft-\modelname &  36.6 &   27.5 & 22.3 &  19.2 \\
        Hard-\modelname  &  \bf{37.4} &  \bf{28.7} & \bf{22.6}  & \bf{19.5}  \\
        \bottomrule
       \end{tabular}
    }
    \caption{BLEU scores on the EN-JA and ZH-EN translation.}
    \label{tab:nmt}
    \vspace{-3mm}
\end{wraptable}
\paragraph{Baselines and Results}
In addition to the conventional seq2seq, we compare with the original SSNT model which only accounts 
for monotonic alignments. We also implemented a variant that combines SSNT with the local reordering module~\cite{huang2017towards} 
as our baseline to show the advantage of our structured ordering module.

Results are shown in Table~\ref{tab:nmt}.
Our model, especially Hard-\modelname, consistently outperforms other baselines on both splits.
In EN-JA translation, the advantage of our best-performance Hard-\modelname is slightly more pronounced in the \textsc{LEN} split than 
in the \textsc{IID} split. In ZH-EN translation, while SSNT and its variant do not outperform seq2seq 
in the \textsc{LEN} split, \modelname can still achieve better results than seq2seq. These results show that our model is better than its alternatives at generalizing to longer sentences for machine translation.

\begin{CJK*}{UTF8}{gbsn}
\begin{table*}[t]
    \center
    \scalebox{0.91}{
    \setlength\tabcolsep{2pt}
   \begin{tabular}{rl}
    \toprule
    original input:  & 在$^1_{\text{in}}$ \enskip 美国$^2_{\text{usa}}$ \enskip 哪些 $^3_{\text{which}}$ \enskip 州 $^4_{\text{state}}$ \enskip 与 $^5$ \enskip 最长$^6_{\text{longest}}$ \enskip 的$^7$ \enskip 河流$^8_{\text{river}}$ \enskip 接壤$^9_{\text{border}}$\\
    reordered input: & 州$^4_{\text{state}}$ \enskip 接壤$^9_{\text{border}}$ \enskip 最长$^6_{\text{longest}}$ \enskip 的$^7$ \enskip 河流$^8_{\text{river}}$ \enskip 与$^5$ \enskip 哪些$^3_{\text{which}}$ \enskip 美国$^2_{\text{usa}}$ \enskip 在$^1_{\text{in}}$  \\
    prediction: &  \<\underline{state}$^4$ \underline{next\_to\_2}$^9$ \underline{longest river}$^{6,7,8}$ \underline{loc\_2 countryid\_ENTITY}$^{5,3,2}$> \\
    ground truth: & \<state next\_to\_2 longest river loc\_2 countryid\_ENTITY> \\
    \hline 
    original input: & according$^1$ \enskip to$^2$ \enskip the$^3$ \enskip newspaper$^4$ \enskip ,$^5$ \enskip there$^6$ \enskip was$^7$ \enskip a$^8$ \enskip big$^9$ \enskip fire $^{10}$ \enskip last$^{11}$ \enskip night$^{12}$  \\
    reordered input:& according$^1$ \enskip to$^2$ \enskip the$^3$ \enskip newspaper$^4$ \enskip ,$^5$ \enskip night$^{12}$ \enskip last$^{11}$ \enskip big$^{9}$ \enskip fire$^{10}$ \enskip  a$^8$   \enskip there$^6$ \enskip was$^7$  \\
    prediction: & \underline{新聞 に よ れ ば 、}$^{1,2,3,4,5}$ \enskip \underline{昨夜}$^{12}$ \enskip \underline{大}$^{11, 9}$ \enskip \underline{火事}$^{10}$ \enskip \underline{が あ}$^{8,6}$ \enskip \underline{っ た}$^{7}$ \\
    ground truth: & 新聞 に よ る と 昨夜 大 火事 が あ っ た \\
    \bottomrule
   \end{tabular} 
}
   \caption{Output examples of Chinese semantic parsing and English-Japanese translation. For clarity, the input words are labeled with position indices, 
   and, for semantic parsing, with English translations. A prediction consists of multiple segments, each annotated with a superscript referring to input tokens.}
  \label{tab:sample-output}
\end{table*}
\end{CJK*}
\paragraph{Interpretability}
Latent alignments, apart from promoting systematic generalization, also lead to better 
interpretability as discrete alignments reveal the internal process for generating output.
For example, in Table~\ref{tab:sample-output}, we show a few   examples from our model. 
Each output segment is associated with an underlying rationale, i.e. a segment of the reordered input.

 \section{Conclusion and Future Work}

 In this work, we propose a new general seq2seq model that accounts for latent segment-level 
 alignments. Central to this model is a novel structured reordering module which is coupled with 
 existing modules to handle non-monotonic segment alignments. We model reorderings as separable permutations and
 propose an efficient dynamic programming algorithm to perform marginal inference and sampling. It allows latent reorderings to be induced with end-to-end training.
 Empirical results on both synthetic and real-world datasets show that our model can achieve
 better systematic generalization than conventional seq2seq models.

The strong inductive bias introduced by modeling alignments in this work could be potentially
beneficial in weakly-supervised and low-resource settings, such as weakly-supervised semantic parsing 
and low-resource machine translation where conventional seq2seq models usually do not perform well. 

\section*{Acknowledgements}
We thank Milo\v{s} Stanojevi\'{c} and Khalil Sima’an for their valuable comments;
Lei Yu and Chris Dyer for providing the preprocessed data for machine translation;
the anonymous reviewers for their helpful feedback.   We gratefully acknowledge the support
of the European Research Council (Titov: ERC
StG BroadSem 678254; Lapata: ERC CoG TransModal 681760) and the Dutch National Science
Foundation (NWO VIDI 639.022.518).

\bibliography{main}
\bibliographystyle{plainnat}

\newpage
\appendix








\section{Appendix}

\subsection{WCFG to PCFG Conversion}

The algorithm of converting a WCFG to its equivalent PCFG is shown in Algorithm~\ref{algo:wcfg-pcfg}. In a bottom-up manner, the algorithm first computes an inner weight $\beta[X_i^k]$ for each segment, which is the total weight of all derivations with root $X_i^k$. Then the algorithm normalizes the weight of production rules whose left-hand side is $X_i^k$ using the inner weight.
The resulting normalized weight for a production rule, e.g., $G[X_i^k \xrightarrow{S} X_i^j X_{j}^k]$, is the conditional probability of applying the rule $X_i^k \xrightarrow{S} X_i^j X_{j}^k$ given the presence of the segment $X_i^k$. The PCFG is equivalent to the original WCFG in the sense that 
for each derivation $D$, we have
\begin{equation*}
    p_\phi(D|x) = \frac{\prod_{R \in D} f_\phi(R)}{Z(x, \phi)} = \prod_{R \in D} G_\phi(R)    
\end{equation*}
where $Z(x, \phi) = \sum_{D'} \prod_{R \in D'} f_\phi(R)$. Full proof of this equivalence can be found in \citet{smith2007weighted}. The factorization of the derivation-level probability to rule-level probability facilitates our design of dynamic programming for marginal inference.  

\begin{algorithm}[h]
    \caption{Converting WCFG to PCFG}
    \begin{algorithmic}[1]
        \State initialize $\beta[\dots]$ to 0
        \For{$i := 0$ \textbf{to} $n-1$} \Comment width-1 spans
            \State $\beta[X_i^{i+1}] = 1$ 
        \EndFor
        \For{$w := 2$ \textbf{to} $n$} \Comment width of spans
            \For{$i := 0$ \textbf{to} $n - w$} \Comment start point
                \State $k := i + w$  \Comment end point
                \For{$j := i+1$ \textbf{to} $k-1$}  \Comment compute inner weight
                   \State $\beta[X_i^k] += f_\phi(X_i^k \xrightarrow{S} X_i^j X_{j}^k)  \beta[X_i^j]\beta[X_{j}^{k}] $ \Comment S: Straight
                   \State $\beta[X_i^k] += f_\phi(X_i^k \xrightarrow{I} X_i^j X_{j}^k) \beta[X_i^j]\beta[X_{j}^{k}] $ \Comment I: Inverted
                \EndFor
                \For{$j := i+1$ \textbf{to} $k-1$} \Comment normalize weight
                   \State $G(X_i^k \xrightarrow{S} X_i^j X_{j}^k)= \frac{f_\phi(X_i^k \xrightarrow{S} X_i^j X_{j}^k) \beta[X_i^j]\beta[X_{j}^{k}]}{\beta[X_i^k]} $
                   \State $G(X_i^k \xrightarrow{I} X_i^j X_{j}^k) = \frac{f_\phi(X_i^k \xrightarrow{I} X_i^j X_{j}^k) \beta[X_i^j]\beta[X_{j}^{k}]}{\beta[X_i^k]} $
                \EndFor
            \EndFor
        \EndFor
        \State \Return $G[\dots]$
    \end{algorithmic}
    \label{algo:wcfg-pcfg}
\end{algorithm}

\subsection{Proof of the Dynamic Programming for Marginal Inference}
We prove the correctness of the dynamic programming algorithm for computing the marginal permutation matrix
of separable permutations by induction as follows.
\begin{proof}
As a base case, each word (i.e., segment with length 1) is associated with an identity permutation matrix $\bs 1$.  
Then we assume that the marginal permutation matrix for all segments with length $1<k-i<n$ is $\mE_i^k$, which is defined 
as $ \E_{p(D_i^k)}[\mM(D_i^k)]$ where $D_i^k$ is the derivation tree of segment $i$ to $k$, and $\mM(D_i^k)$ 
is the permutation matrix corresponding to $D_i^k$.
It is obvious that $\mE_i^{i+1} =  \bs 1$.
The marginal permutation matrix for all segments with length $n$ can be obtained by 
\begin{align*}
\mE_i^k & = \E_{p(D_i^k)}[\mM(D_i^k)] \\
&= \sum_{i<j<k} \Big( G_\phi(\mathcal{S}_{i,j,k}) \big(\E_{p(D_i^j)}[\mM(D_i^j)]) \oplus \E_{p(D_j^k)}[\mM(D_j^k)] \big) \\
& \quad \quad + G_\phi(\mathcal{I}_{i,j,k}) \big(\E_{p(D_i^j)}[\mM(D_i^j)] \ominus \E_{p(D_j^k)}[\mM(D_j^k)]\big) \Big) \\
&= \sum_{i<j<k} \Big ( G_\phi(\mathcal{S}_{i,j,k}) (\mE_i^j \oplus \mE_j^k) + G_\phi(\mathcal{I}_{i,j,k}) (\mE_i^j \ominus \mE_j^k) \Big ) \\
\end{align*}
where in the second step we consider all the possible expansions of the derivation tree $D_i^k$;
in the third step, we obtain the recursion that is used in Step 12-14 of Algorithm 1 by reusing the 
marginal permutations matrices of shorter segments.
\end{proof}

\subsection{Architecture and Hyperparameters}
\begin{figure}[t]
    \centering
   \includegraphics[width=1.0\textwidth]{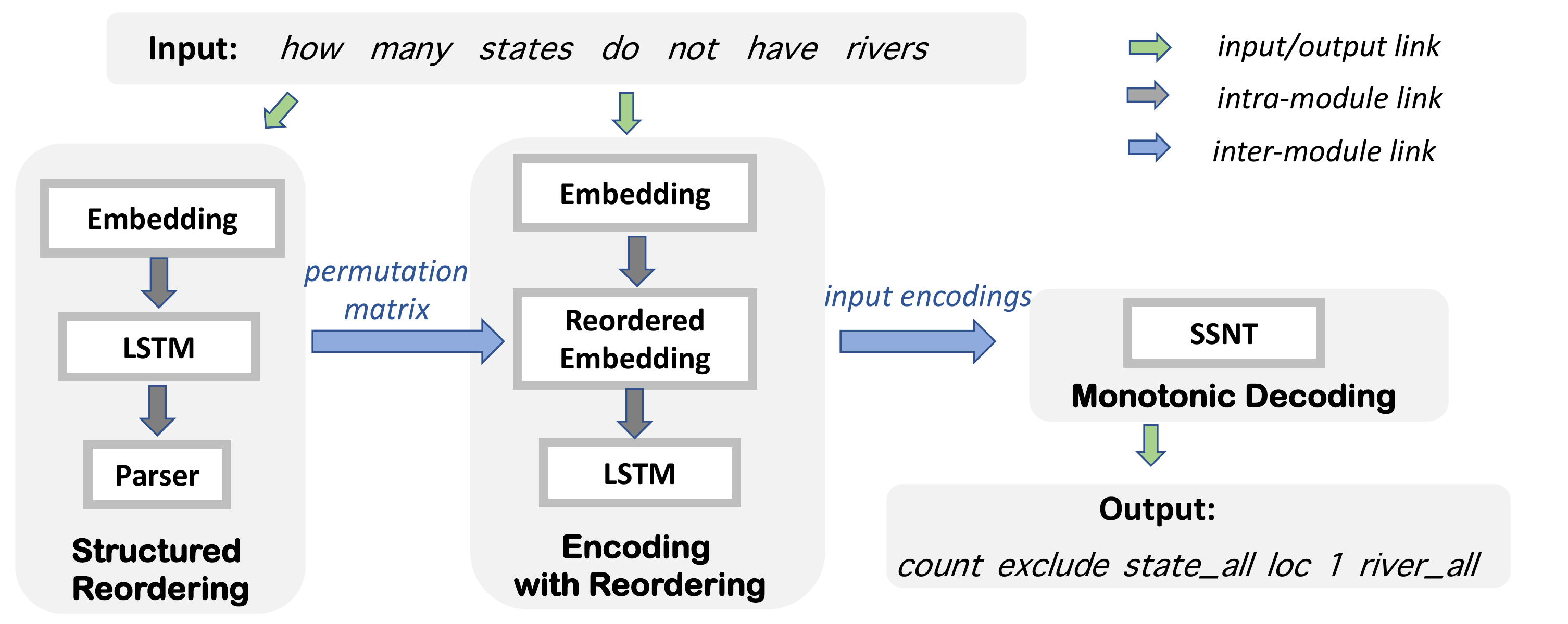} 
   \caption{The detailed architecture of our seq2seq model for semantic parsing (view in color). 
   First, the structured reordering module genearates a (relaxed) permutation matrix given the input utterrance. 
   Then, the encoding module generates the representations of the input utterance based on the reordered embeddings, which
   are computed based on the original embedding and the permutation matrix computed in the first step.  
   Finally, the decoding module, namely SSNT, generates the output program monotonically based on the input encodings.}
   \label{fig:nn-arch}
\end{figure}
\begin{table}[t]
    \center
    \scalebox{1.0}{
    \begin{tabular}{cc} 
        \toprule
        Name & Range \\
        \midrule
        embedding size & [128, 256, 512] \\
        number of encoder LSTM layer & [1,2] \\
        encoder LSTM hidden size & [128, 256, 512] \\
        decoder LSTM layer & [1,2] \\
        decoder LSTM hidden size & [128, 256, 512] \\
        decoder dropout & [0.1, 0.3, 0.5, 0.7, 0.9] \\
        temperature of Gumbel-softmax & [0.1, 1, 2, 10] \\ 
        label smoothing & [0.0, 0.1] \\
        \bottomrule
       \end{tabular}
    }
    \caption{Main hyperparameters of \modelname.}
    \label{tab:hyperparameters}
\vspace{-3mm}
\end{table}

The detailed architecture of \modelname is shown in Figure~\ref{fig:nn-arch}.
In the structured reordering module, we compute the scores for BTG production rules  
using span embeddings~\cite{wang2016graph} followed by a multi-layer perceptron.
Specifically, the score function for each rule has form $G(R_{i,j,k})=\text{MLP}(\vs_{ij}, \vs_{jk})$, 
where $\vs_{ij}$ and $\vs_{jk}$ are the span embeddings based on \citep{wang2016graph}, $\text{MLP}$ is 
a multi-layer perceptron that outputs a 2-d vector, which corresponds to the score of $R$=Straight and $R$=Inverted, respectively.
Similar to a conventional LSTM-based encoder-decoder model, 
LSTMs used in structured reordering and encoding module are bidirectional whereas 
the LSTM for decoding (within SSNT) is unidirectional.
We implemented all models using Pytorch~\cite{paszke2019pytorch}. 
We list the main hyperparameters we tuned are shown in Table~\ref{tab:hyperparameters}.
The full hyperparameters for each experiment will be released along with the code.

\subsection{Training Strategy}
Empirically, we found that during training the structured reordering module tends to converge 
to a sub-optimal point where it develops a simple reordering strategy and the subsequent modules
(i.e., the encoding and decoding module in Figure~\ref{fig:nn-arch}) quickly 
adapt to naive reorderings.  For example, in the EN-JA translation task,
the reordering module tends to completely invert the input English translation after training.
While this simple strategy proves to be a useful heuristic~\citep{katz2008syntactic}, we would like more 
accurate reordering to emerge during training.
This issue is similar to posterior collapse~\citep{bowman2015generating}, 
a common issue in training variational autoencoders.

Inspired by \citet{he2019lagging}, we speculate that the issue occurred due to 
that optimization of the structured reordering module usually lags far behind 
the optimization of subsequent modules during the initial stages of training.
We use a simple training strategy to alleviate the issue.
Specifically, during the initial $M$ training steps, with a certain probability $p$, we only update 
the  parameters of the structured reordering module and ignore the gradients of the parameters 
from the subsequence modules. 
$M$ and $p$ are treated as hyperparameters.
With this strategy, the structured reordering module is updated more often than the subsequent modules, and
has a better chance to catch up with the optimization of subsequent modules.  
We find that this simple training strategy usually leads to better segment alignments and better performance.




\end{document}